\documentclass[preprints,article,accept,moreauthors,pdftex]{Definitions/mdpi}

\firstpage{1} 
\pubvolume{xx}
\issuenum{1}
\articlenumber{5}
\pubyear{2019}
\copyrightyear{2019}
\history{}

\Title{Analysis of Explainers of Black Box Deep Neural Networks for Computer Vision: A Survey}

\Author{Vanessa Buhrmester $^{*}$\orcidA{}, David M\"unch \orcidB{}, and Michael Arens \orcidB{} }

\AuthorNames{Vanessa Buhrmester, David M\"unch and Michael Arens}

\address{Fraunhofer IOSB, Gutleuthausstra{\ss}e 1, 76275 Ettlingen, Germany}

\corres{Correspondence: vanessa.buhrmester@iosb.fraunhofer.de; Tel.: +49 7243-992-167}

\abstract{Deep Learning is a state-of-the-art technique to make inference on extensive or complex data. As a black box model due to their multilayer nonlinear structure, Deep Neural Networks are often criticized to be non-transparent and their predictions not traceable by humans. Furthermore, the models learn from artificial datasets, often with bias or contaminated discriminating content. Through their increased distribution, decision-making algorithms can contribute promoting prejudge and unfairness which is not easy to notice due to lack of transparency.
Hence, scientists developed several so-called explanators or explainers which try to point out the connection between input and output to represent in a simplified way the inner structure of machine learning black boxes. In this survey we differ the mechanisms and properties of explaining systems for Deep Neural Networks for Computer Vision tasks. We give a comprehensive overview about taxonomy of related studies and compare several survey papers that deal with explainability in general. We work out the drawbacks and gaps and summarize further research ideas.}


\begin{document}


\section{INTRODUCTION}
\label{sec:intro}
Artificial Intelligence (AI)-based technologies are increasingly being used to make inference on classification or regression problems: Automated image and text interpretation in medicine, insurance, advertisement, public video surveillance, job applications, or credit scoring save staff and time and are moreover practical successful. The severe drawback is that many of these technologies are black boxes and referenced results can hardly be understood by the user. 

\subsection{Motivation: More complex algorithms are hardly comprehensible.}
\label{sec:motivation}
\begin{figure}[H]
	\begin{center}
\includegraphics[width=0.4\linewidth]{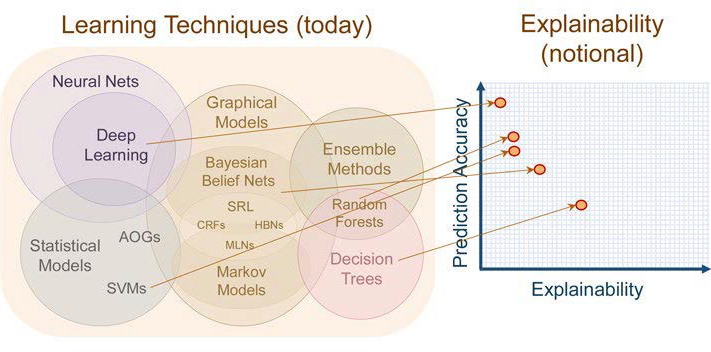}
\end{center}
	\caption{The explainability is decreasing with a growing prediction accuracy while the prediction accuracy is growing with more complex models like Deep Learning, Figure from \cite{gunning2017} .}
	\label{fig:complex}
\end{figure}

Latest models are more complex and Deep Learning (DL) architectures are getting deeper and deeper and millions of parameters are calculated and optimized by a machine. For example, the common network VGG-19 incorporates about 144 millions parameters that were optimized over millions or hundred thousands of images \cite{simonyan2014very}. ResNet has about $5\cdot 10^7$ trainable parameters and for classifying one image it needs to execute about $10^{10}$ floating point operations, see \cite{he2016deep}. It is hardly traceable and not recalculate-able by humans. Metrics like accuracy or the mean average precision are depending on the quality of manually hand annotated data. However, these metrics are often the only values that evaluate the learning algorithm itself.
The explainability of a Machine Learning (ML) technique is decreasing with an increasing prediction accuracy, and the prediction accuracy is growing with more complex models like Deep Neural Networks (DNNs), see Figure \ref{fig:complex}. But a trade-off between explainability and accuracy is not satisfying, especially in critical tasks.

\begin{figure}[H]
	\begin{center}
\includegraphics[width=0.4\linewidth]{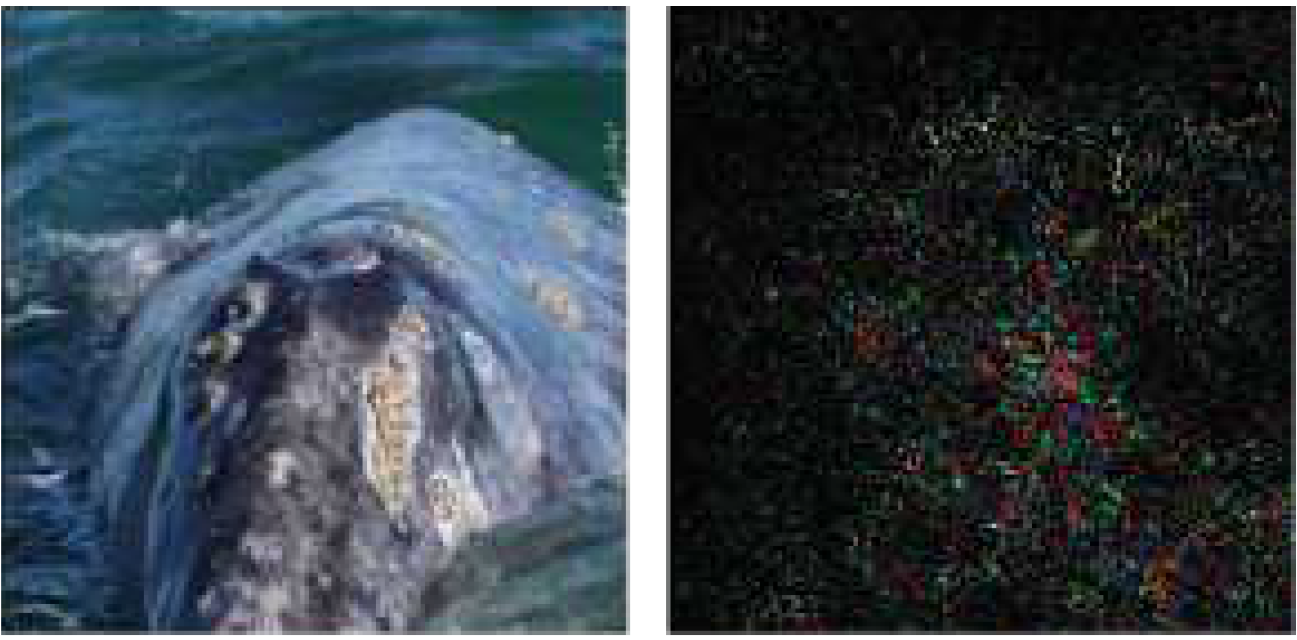}
\end{center}
	\caption{\textbf{DeepFool} \cite{deepfoool2016} examines the robustness of neural networks. Very small noisy images were added on right classified images, humans cannot see the difference, but the algorithms changes its prediction: $x$ (left) is correctly classified as whale, but $x+r$ as turtle, $r$ (right) is very small.
}
\label{fig:deepfool}
\end{figure}

\begin{figure}[H]
	\begin{center}
\includegraphics[width=0.4\linewidth]{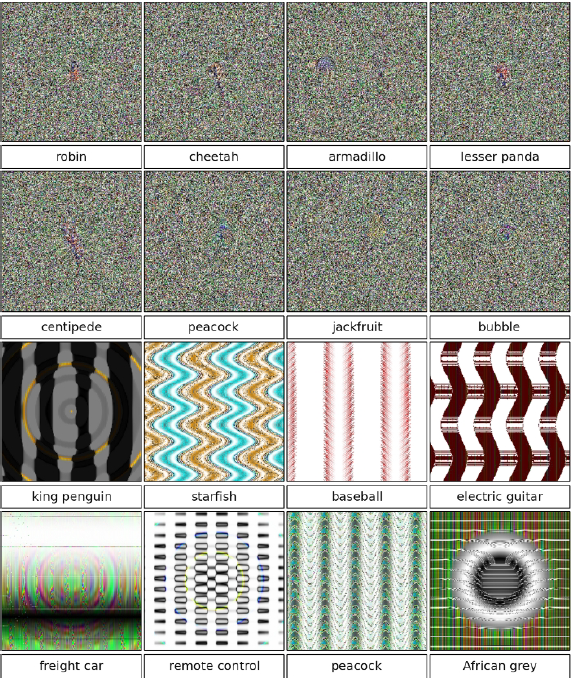}
\end{center}
	\caption{ \cite{nguyen2015} created artificial images, that where unrecognizable by humans, but the state-of-the-art classifier was very confident that they were known objects. Images are either directly (top) or indirectly (bottom) encoded. }
	\label{fig:nguyen}
\end{figure}

\begin{figure}[H]
	\begin{center}
\includegraphics[width=0.4\linewidth]{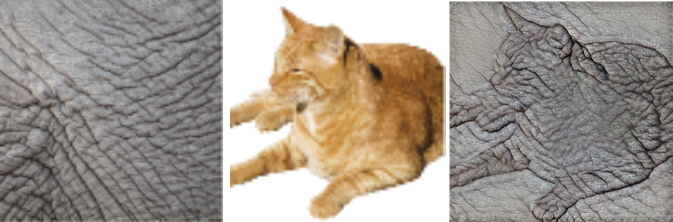}
\end{center}
	\caption{Texture-shape cue conflict \cite{geirhos2018imagenet}: texture (left) is classified as elephant, content (middle) is classified as cat, texture-shape (right) is classified as elephant because of a texture bias.}
	\label{fig:geirhos1}
\end{figure}

\begin{figure}[H]
	\begin{center}
\includegraphics[width=0.6\linewidth]{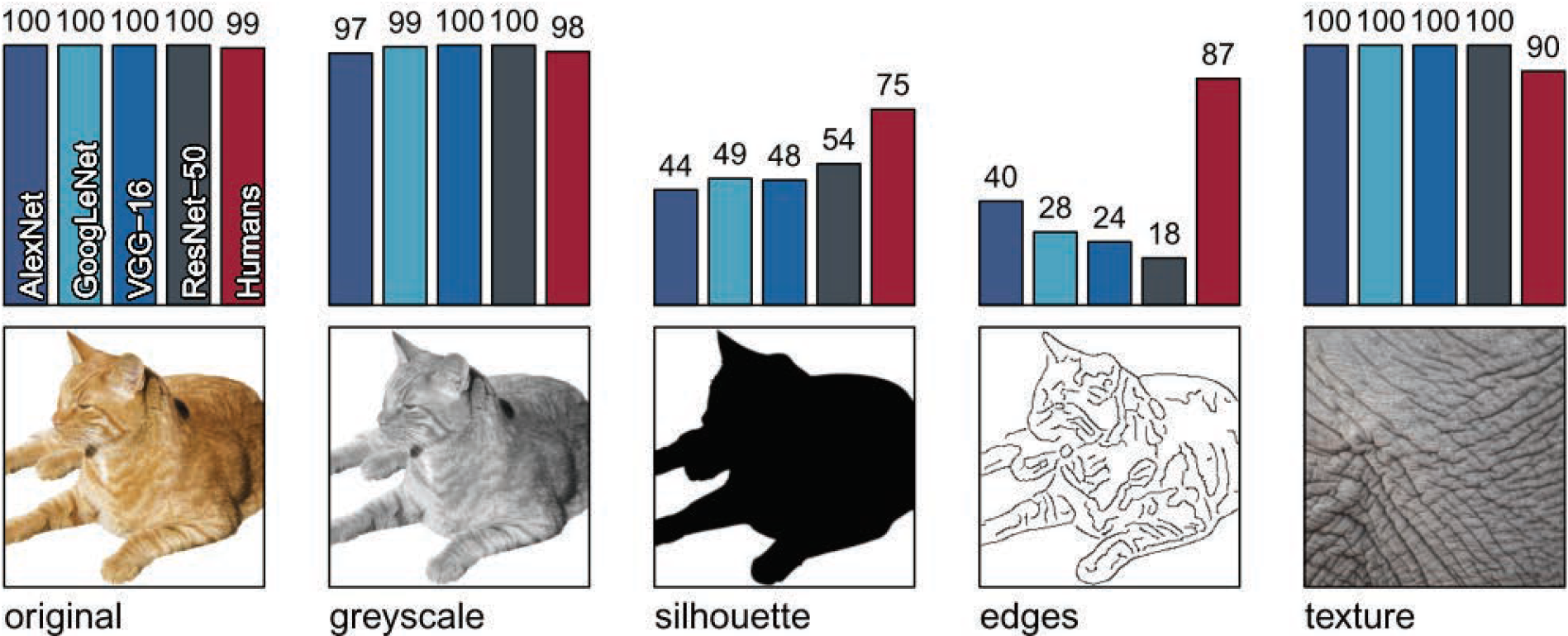}
\end{center}
	\caption{\cite{geirhos2018imagenet} show that Convolutional Neural Networks focus more on texture than on shapes and that it is a good idea to improve shape bias to get more reliable results in the way like humans would interpret the content of an image.}
	\label{fig:geirhos2}
\end{figure}

 In the past, many scientists found weaknesses of Deep Learning models -- even high performing models are affected. They showed, for instance, how easy one could fool object detectors with small changes in the input image or created adversarial examples to make them collapse, see \cite{szegedy2013intriguing}, \cite{goodfellow2014explaining}. Furthermore, they draw attention on supposedly good models which focus on totally wrong features and just made good decisions by chance. The problem is that the neural network learns only from the training data, which should characterize the task. But suitable training data is tedious to create and annotate, hence it is not always perfect. If in the training data is a bias, the algorithm will learn it as well. The following examples present attacks on neural networks:

A classifier of enemy tanks and friendly tanks with high accuracy did not deliver good results in the application and was discovered to be a just a good classifier of sunny or overcast days \cite{freitas2014}. The reason was that most of the photos for the training set of enemy tanks where taken on days with clouds on the sky, while the friendly ones were shot during sunny weather. The same problem happened as a dog-or-wolf-classifier turned out to be just a good snow detector \cite{ribeiro2016should} because of a bias in the background of the training images.
There are several cases which underline the negative characteristics of a DNN. Changing only one pixel in an input image or one letter in a sentence could change the prediction, or even adding small-magnitude perturbations, \cite{deepfoool2016}, \cite{bose2018adversarial}, \cite{jia2017adversarial}, see also Figure \ref{fig:deepfool}. Adversarial examples with serious impact exist; fixed stickers on road signs \cite{eykholt2018robust} or extended adversarial patch attacks in optical flow networks \cite{flowattack} could lead to dangerous misinterpretation. Worn prepared glasses confuse face detectors  by imitating special persons, \cite{sharif2016accessorize}, \cite{sharif2019general}. Further cases regard on Figure \ref{fig:nguyen}, \ref{fig:geirhos1}, and \ref{fig:geirhos2}. All this examples show how harmful it could be to rely on a black box with supposedly well performing results. However, currently applied DNN-methods and models are such vulnerable black boxes.


\subsection{Also an ethical question: Why is explainability so important?}
\label{sec:ethics}

Machine Learning-models are tested by the engineer on validation data, this is the basis on which he evaluates how correct they work according to the chosen metrics. Although, the applicant applies the model to the real world, he should believe that the results still are reliable. But how could he? However, fairness and reliability should be important properties of such models. To win the users trust, the link between the features and the predictions should be understandable. If a human is even judged by a machine, he or she should has the right of explanation.
Since the European Union’s new General Data Protection Regulation (in Germany DSGVO) was passed in the EU in May 2018, it will restrict the use of machine learning and automated individual decision-making and focus on protection of  sensitive dates of persons like their age, sex, ancestry, name or place of residence for instance. If a result affects users, they should be able to demand for explanations of the algorithmic decision that was made about them, see \cite{goodman2016}. For example, if a doctor makes a mistake the patient wants to know why. Was the mistake excusable? Or does the mistake reflect negligence, even intentionally or occurred due to other factors?
Similarly, if an algorithm fails and contributes to an adverse clinical event or malpractice, doctors need to be able to understand why it produced the result and how it reached a serious decision.

Although or just because a human decision is not free from prejudice, it must be ensured that an algorithm for selection procedure -- e.g. for job proposals or suspended sentences -- may not discriminate humans because of sex or origin etc. Disadvantages can arise here for individuals, groups of persons or a whole society. The prevailing conditions can thereby further deteriorate more and more caused by a gender-word bias \cite{holmes2008handbook}. An interesting example are word embeddings \cite{bolukbasi2016man} that a Natural Language Processing (NLP) algorithm creates from training datasets, which are just any available texts of a special origin \cite{hirschberg2015advances}. The procurement of women, disabled, black people etc. seem to be deep anchored in these texts, with the serious consequence that the model learns that to be currently. This reinforces discrimination and unfairness. Implicit Association Tests \cite{greenwald1998measuring} have uncovered stereotype biases which people are not aware of. If the model supposes -- and studies \cite{bolukbasi2016man}, \cite{chakrabortyreducing}, \cite{font2019equalizing} show that -- that doctors are male and nurses are female, furthermore, women are sensitive and men are successful, it will sort out all women who apply as a chief doctor -- only because of their sex -- no need to check their qualification. If in the train data foreigners have predominantly less income and increased unemployment, an automatic credit scoring model will suggest a higher interest rate or even refuse the request only because of the origin and without considering the individual financial situation. Unfairness will progress through these algorithms.

Hence, getting understanding and insights in the mechanisms should uncover these problems. Properties of a model like transparency and interpretability are basics to build patient, provider trust, and fairness. If this succeeds, the causes of the discrimination and serious mistakes can be remedied, additionally. There is the opportunity to improve the society through making automated decisions free of prejudice. Our contribution on the way achieving this goal is giving an overview about state-of-the-art-explainers with regard to their taxonomy through differing their mechanisms and technical properties. We do not just limit our work to explaining methods, but also look at the meaning of understanding a machine in general.  To our knowledge this is the first survey paper that focuses on ML-black box DNNs for Computer Vision tasks.

\section{Overview about explaining systems of DNNs}
\label{sec:explainers}
We just give a short introduction in early approaches to explain inner Machine Learning operations. After that we will focus only on understanding DNNs.

\subsection{Early Machine Learning explaining systems}
\label{sec:early}

Early explaining systems for ML-black boxes go back to 1986 with generalized addictive models (\textbf{GAM}) \cite{hastie1986}. This is a global statistic model which uses smooth functions as a diagnostic tool. 
Later \textbf{Decision Trees} were successful classification tools that provide individual explanations, see \cite{craven1996extracting}, \cite{friedman2001elements}. A Decision Tree is a tree-like graph of decisions and their possible consequences that visualizes an algorithm that only contains condition control statements.
Another approach \cite{friedman2001greedy} shows the marginal effect of one or two features on the prediction of learning techniques using Partial Dependence Plots (\textbf{PDP}). The method gives a statement about the global relationship of a feature and whether its relation to the outcome is linear, monotonous or more complex.
PDP is the average of Individual Conditional Expectation (\textbf{ICE}) over all features. ICE \cite{goldstein2015peeking} points to how the prediction changes if a feature changes. PDP is limited to two features.
Another example is the early use of explainable AI \cite{van2004explainable} which was developed as a simulator game for the commercial platform training aid Full Spectrum Command motivated by previous work such as \cite{shortliffe1977mycin}. 
The proposed procedure of \cite{baehrens2010explain} based on a set of assumptions, which allows to explain the decision for a particular label of a single data instance for several classification models. The framework provides local explanation vectors as class probability gradients which yield the relevant features of every point of interest in the data space.\\
We do not respond further to early studies of explaining systems of \textbf{Random Forests} \cite{breiman2001random}, \textbf{Na{\"i}ve Bayesian} classifiers \cite{kononenko1993inductive}, \cite{becker2001visualizing}, \cite{movzina2004nomograms}, Support Vector Machines (\textbf{SVMs}) \cite{poulet2004svm}, \cite{hamel2006visualization}, or other early Machine Learning prediction methods \cite{breiman1996bagging}.

\subsection{Mechanism and properties of DNN-explainers}
\label{sec:mech}

\begin{figure}[H]
	\begin{center}
\includegraphics[width=0.6\linewidth]{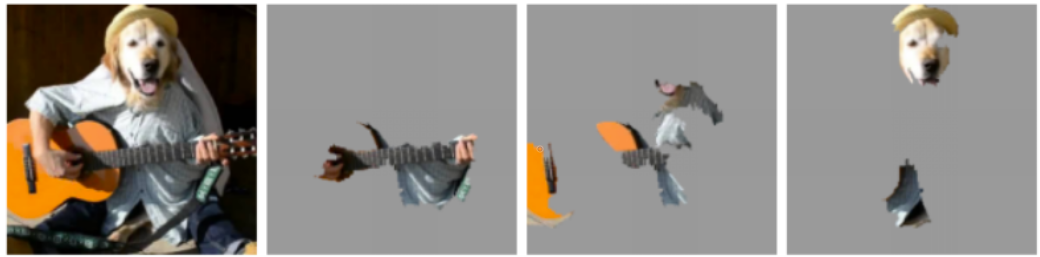}
\end{center}
	\caption{\cite{ribeiro2016should} presented Local Interpretable Model-agnostic Explanations (LIME) which can explain the predictions of any agnostic black box classifier and any data. Here the superpixels -- areas of an input image -- are highlighted that are most responsible for top three image classification predictions. (1) original image (2) explaining electric guitar, (3) explaining acoustic guitar, and (4) explaining labrador.}
	\label{fig:lime}
\end{figure}

In the last years, the importance of DNNs in inference tasks grew rapidly, and with the increasing complexity of these models, also the need of better explanations did. Most used DNNs for image or video processing \cite{ciresan2012deep}, \cite{ciregan2012multi},~ \cite{le2011learning} are Convolutional Neural Networks (CNNs) \cite{krizhevsky2012imagenet}, for videos \cite{le2011learning} or sequences of text Recurrent Neural Networks (RNNs) \cite{rumelhart1988learning}, and especially for language modeling \cite{collobert2011natural} Long-Short Term Memories (LSTMs) \cite{hochreiter1997long}. There exist some general surveys of methods for explaining several machine learning black boxes, so called eXplanatory Artificial Intelligence (XAI), that cover a wide spectrum of AI-based black boxes, for instance see \cite{guidotti2019survey}, \cite{goebel2018explainable}. However, we just want to focus on the black box DNN and deepen the insights, especially in Computer Vision. In the following we describe state-of-the-art explainers in these tasks. There are explainers that create visualizations which give an explanation by digesting the mechanisms of a model down to images which themselves have to be interpreted. Salience maps of important features are calculated, for example, they show super-pixels or given keywords that have influenced the prediction most, see \cite{ribeiro2016should} and Figure \ref{fig:lime}. The problem is that these explainers are in turn black boxes. White box explainers use methods that gain insights and show all causal effects, for instance linear regression or Decision Trees.
Black box explainers do not require access to the internals and do not disclose all feature interaction. There are mainly two kinds of explaining models:
\begin{itemize}
\item \textbf{Ante-hoc} or \textbf{intrinsically interpretable} models, see \cite{lipton2016mythos}. Ante-hoc systems give explanations starting from the beginning of the model. For instance enabling one to gauge how certain a neural network is about its predictions. 
\item \textbf{Post-hoc} techniques entail baking explainability into a model from its outcome, like marking what part of the input data is responsible to the final decision, for example in LIME. These methods can be applied more easily to different models, but say less about the whole model in general.
\end{itemize}
They can be also split in 
\begin{itemize}
\item \textbf{local}: the algorithm can be explained only for each single prediction and
\item \textbf{global}: the whole system can be explained and the logic can be followed from the input to every possible outcome, 
\end{itemize}and even in \begin{itemize}
\item\textbf{model specific}, tied to a particular type of black box or data and
\item \textbf{model agnostic}, indifferently usable.
\end{itemize}

\noindent \cite{gilpin2018explaining} defined key terms like “explanation”, “interpretability”, and “explainability” philosophically. Important is that interpretability and explainability is not the same, although it is often used interchangeably. An explanation is much more concrete, a coherence of facts can be describe with words while an interpretation is just a substantial formation that arises in the head. They also describe the difficulties of both interpretability and completeness, so a compromise is needed. For more details see the following. We compile the most important definitions of properties of explainers and their obvious connections in a more technical way: 

\begin{itemize}
\item An \textbf{explainer} or \textbf{explanator} is a synonym for an explaining system, that gives an answer to a questions. For instance if the question is how a machine is working, the explainer makes the internal structure of a machine more transparent to humans. A further question could be why a prediction was made instead of another, so the explainer should point to where the decision boundaries between classes are and why particular labels are predicted for different data points \cite{ridgeway1998interpretable}.
\item \textbf{Interpretability} is a substantial first step to reach a comprehension of a complex coherence in some level of detail but is insufficient alone, see \cite{gilpin2018explaining}.
\item \textbf{Explainability} includes interpretability but not always reverse. It provides relevant responses to questions and subdivides their meaning in understandable terms to a human, see \cite{gilpin2018explaining}, \cite{doshi2017towards}.
\item \textbf{Comprehensibility} or \textbf{understandable explanation}. An understandable explanation must be created by a machine in a given time and can be comprehend by a user, who need not to be an expert, but has an educational background, in a given time (e.g. one hour or one day).
\item \textbf{Completeness}. A \textbf{complete explanation} records all possible factors from input to output of a model. A DNN with its millions of parameters is too complex, hence a complete explanation would not be understandable. That makes it necessary to focus on the most important reasons and not all of them.
\item \textbf{Compactness}. A \textbf{compact explanation} has a finite number of aspects. 
Because the parameters and operations of a DNN are finite, one can get a complete explanation of a DNN after a finite number of steps. That is why compactness follows from completeness if the regarded connections are finite. A DNN can be explained for instance completely and compactly or compactly and understandably. 
\end{itemize}

\begin{figure}[H]
	\begin{center}
\includegraphics[width=0.6\linewidth]{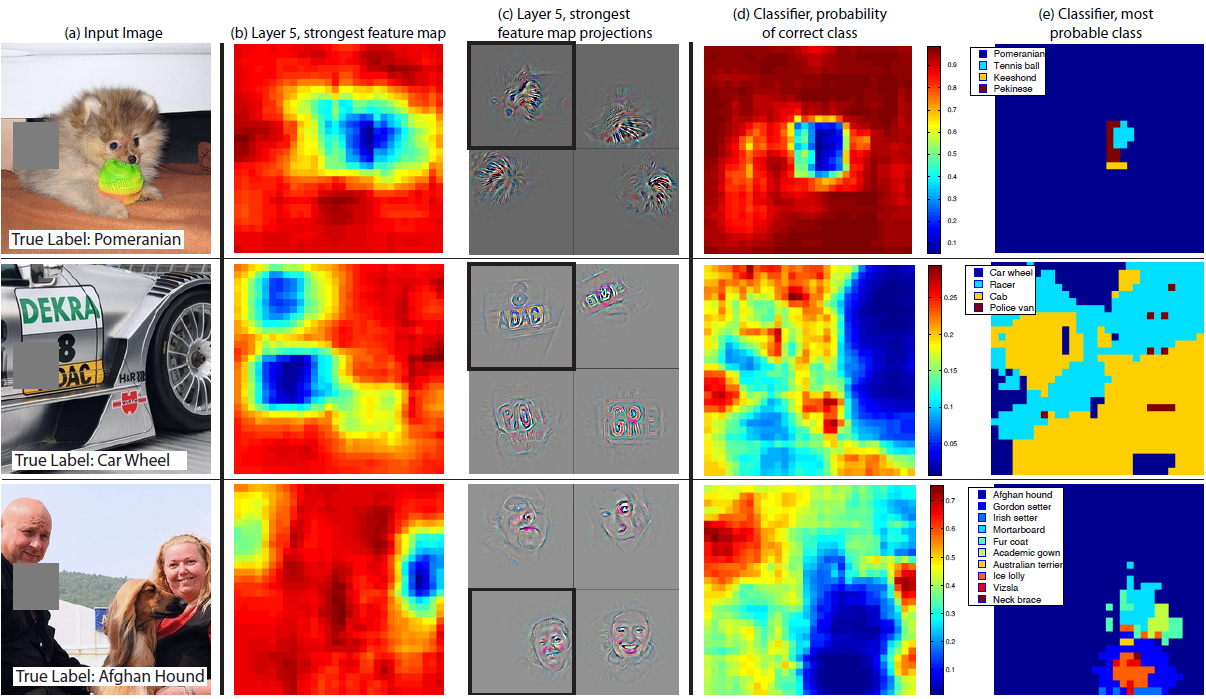}
\end{center}
	\caption{\textbf{Deconvnet} \cite{zeiler2014visualizing}: They plot three examples of input image (a), strongest feature map (2) and feature map projections (3) of layer 5, and the classifier with the probability of correct class (4) and most probable class (5), respectively.}
	\label{fig:zeiler}
\end{figure}

\noindent Other, in our eyes less gentle aspects that are mentioned in the literature are fidelity, trust, intelligibility, privacy, usability, monotonicity, causality, scalability, and generality, see \cite{alexandrov2017explainable}, \cite{doshi2017towards}, \cite{guidotti2019survey}. Now we investigate the employed mechanisms of explainers:
\begin{itemize}
\item \textbf{Visualizations}. To visualize an explanation there are many options \cite{samek2016evaluating}. One tool is to look at the \textbf{activations} produced on each layer of a trained CNN as it processes an image or video. Another one enables visualizing features at each layer of a DNN via regularized optimization in image space \cite{yosinski2015understanding}. Visualizations of particular neurons or neuron layers show responsible features that lead to a maximum activation or highest possible probability of a prediction, see \cite{dosovitskiy2016inverting} and can be split in \textbf{generative models} or \textbf{saliency maps}.  To create a map of import pixels one can repeatedly feed an architecture with several portions of inputs and compare the respective output. Or one visualizes them directly by going rearwards through the inverted network from an output of interest. Also grouped in this category is exploiting neural networks with \textbf{activation atlases} through feature inversion. This method can reveal how the network typically represents some concepts \cite{carter2019activation}. 
\item \textbf{Gradients} or variants of (guided) \textbf{backpropagation} can emphasize important unit changes and thereby draw attention to sensitive features or input data areas, see \cite{springenberg2015striving}, \cite{zhang2016colorful}, \cite{zhang2018top}.  With these techniques it is also able to produce artificial prototype class member images that maximize a neuron activation or class confidence value, see \cite{nguyen2016synthesizing}, \cite{simonyan2013deep}. 
\item Regarding image or text portions that \textbf{maximize the activation of interesting neurons or whole layers} can lead to interpretation of the responsible area of individual parts of the architecture.
\item \textbf{Deconvolution} or \textbf{inverting DNNs} is applied to create typical inputs or parts of an input, that fits to a desired output of the network, a special layer or single unit, see \cite{zeiler2014visualizing}, Figure \ref{fig:zeiler}, \cite{mahendran2015understanding}. 
\item Another method is \textbf{decomposition}, isolation, transfer or limitation of portions of networks, e.g. layers to get further insights in which way single parts of the architecture influences the results, see \cite{bach2015pixel}, or Deep Taylor Decomposition (\textbf{DTD}), \cite{montavon2017explaining}. \textbf{Automatic Rule Extraction} and \textbf{Decision Trees} are anchored in this area, too. \end{itemize}

We give some examples of explaining approaches that can be placed in the last point: A global ante-hoc method for tabulator data is Bayesian Rule List (\textbf{BRL}) \cite{letham2015interpretable}, \cite{ustun2014methods}. BRL is a generative model that yields a posterior distribution over possible decision lists which consist of a series of if-then-statements. The ``if"-statements define a partition of a set of features and the ``then"-statements
correspond to the predicted outcome of interest. The work was developed from preliminary versions that used a different prior, called Bayesian List Machine \cite{letham2013interpretable}. Similar to \textbf{DeepRED} \cite{zilke2016deepred} the rule generation ``KnowledgeTron" (\textbf{KT}) \cite{fu1994rule} applied an if-then-rule for each neuron, layer by layer. Another option in this field is to decompose a DNN in \textbf{Decision Trees}, e.g. DeepRED or \cite{friedman2001elements}. Decision trees were used since the 1990s to explain machine learning tasks, but applied to DNNs their generation is quite expensive and the comprehensibility suffers from the necessary size and number of trees. Decision Trees are able to explain an algorithm completely but with DNNs there is a conflict with comprehensibility.  

\subsection{Selected DNN-explainers presented}
\label{sec:selected}
Let us give a technical overview about selected explanator models, we will put a focus on the category Computer Vision:

\noindent The Counterfactual Impact Evaluation (\textbf{CIE}) method, \cite{bottou2013counterfactual}, \cite{hainmueller2014kernel},  is a local method of comparison for different predictions. Counterfactuals are contrastive. They explain why a decision was made instead of another. A counterfactual explanation of a prediction may be defined as the smallest change to the feature values that changes the prediction to a predefined output. They could be employed to DNNs with any data type.

Famous work to visualize and understand Convolutional Neural Networks is \textbf{Deconvnet} \cite{zeiler2014visualizing}: Deconvnet is a calculation of a backward convolutional network that reuses the weights at each layer from the output layer back to the input image. The employed mechanisms were deconvolution and unpooling which are especially designed for CNNs with convolutions, maxpooling, and REctified Linear Units (ReLUs). The method makes it possible to create feature maps of an input image that activates certain hidden units most, linked to a particular prediction, see Figure \ref{fig:zeiler}. With their propagation technique they identify the most responsible patterns for this output. The patterns are visualized in the input space. Deconvnet is limited to max-pooling layers and in the absence of a particular theoretical criterion which could directly connect the prediction to the created input patterns. To close that gap \cite{springenberg2015striving} proposes a new and efficient way following the initial attempt of Zeiler and Fergus. They replaced max-pooling by a convolutional layer with increased stride and called the method \textbf{The All Convolutional Net}. The performance on image recognition benchmarks was similar well. With this approach they were able to analyze the neural network by introducing a novel variant of the Doconvnet to visualize the concepts learned by higher network layers of the CNN. The problem of max-pooling layers is that they are not invertible in general. That is the reason why Zeiler and Fergus computed positions of maxima within each pooling region and used these ``switches" in the Deconvnet for a discriminative reconstruction. Not using max-pooling Springenberg et al. could directly display learned features and was not conditioned on an image. Furthermore, for higher layers, they produced sharper, more recognizable visualizations of descriptive image regions than previous methods. This is in agreement with the fact that higher layers learn more invariant representations. 

\cite{yosinski2015understanding} introduced two tools to aid interpreting DNNs in a global way.
First they have displayed the neurons activations produced on each layer of a trained CNN processing an image or sequence of images. They found that looking at live activations that change in response to input images helps to build valuable intuitions about the inner mechanisms of these neural networks. The second tool was built on previous versions which calculated less recognizable images. Some novel regularization methods that were combined produce qualitatively clearer and more interpretable visualizations and enable plotting features at each layer via regularized optimization in image space.

\begin{figure}[H]
	\begin{center}
\includegraphics[width=0.4\linewidth]{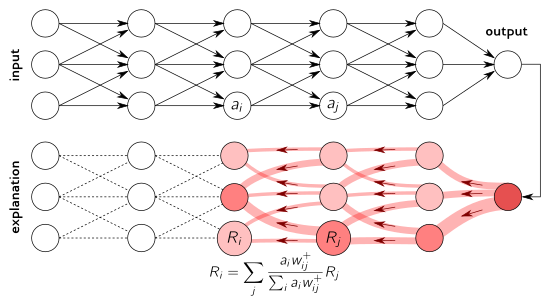}
\end{center}
	\caption{Layer-wise Relevance Propagation (LRP) \cite{bach2015pixel} is a gradient method suffering from shattered gradient problems. The idea behind is a decomposition of the prediction function as a sum of layer-wise relevance values. When LRP is applied to deep ReLU networks, LRP can be understood as a deep Taylor decomposition of the prediction.}
	\label{fig:lrp}
\end{figure}

Gradient-based is layer-wise Relevance Propagation (\textbf{LRP}) \cite{bach2015pixel} which is suffering from shattered gradient problems, see Figure \ref{fig:lrp}. It relies on a conservation principle to propagate the outcome decision back without using gradients. The idea behind is a decomposition of prediction function as a sum of layer-wise relevance values. When LRP is applied to deep ReLU networks, LRP can be understood as a deep Taylor decomposition of the prediction. This principle ensures that the prediction activity is
fully redistributed through all the layers onto the input variables. More about how to explain nonlinear classification decisions with Deep Taylor Decomposition see \cite{montavon2017explaining}. They decompose the network classification decision into contributions of its input elements and assess the importance of single pixels in image classification tasks. Their method efficiently utilizes the structure of the network by backpropagating the explanations from the output to the input layer and display the connections in heat maps. 

Also a global and ante-hoc model is a joint framework for description and prediction, presented by \cite{lakkaraju2016interpretable}. The model creates Black box Explanations through Transparent Approximations (\textbf{BETA}). It learns a compact two-level decision set in which each rule explains parts of the model behavior unambiguously and is a combined objective function to optimize these aspects: high agreement between explanation and the model, little overlaps between decision rules in the explanation, and the explanation decision set is lightweight and small.

An interpretable end-to-end explainer for healthcare is the REverse Time AttentIoN mechanism \textbf{RETAIN} \cite{choi2016retain} for application to Electronic Health Records (EHR) data. The approach mimics physician practice by attending the EHR data, two RNNs are trained in a reverse time order with the goal of efficiently generating the appropriate attention variables. It is based on a two-level neural attention generation process that detects influential past visits and significant clinical variables to improve accuracy and interpretability.

\begin{figure}[H]
	\begin{center}
\includegraphics[width=0.6\linewidth]{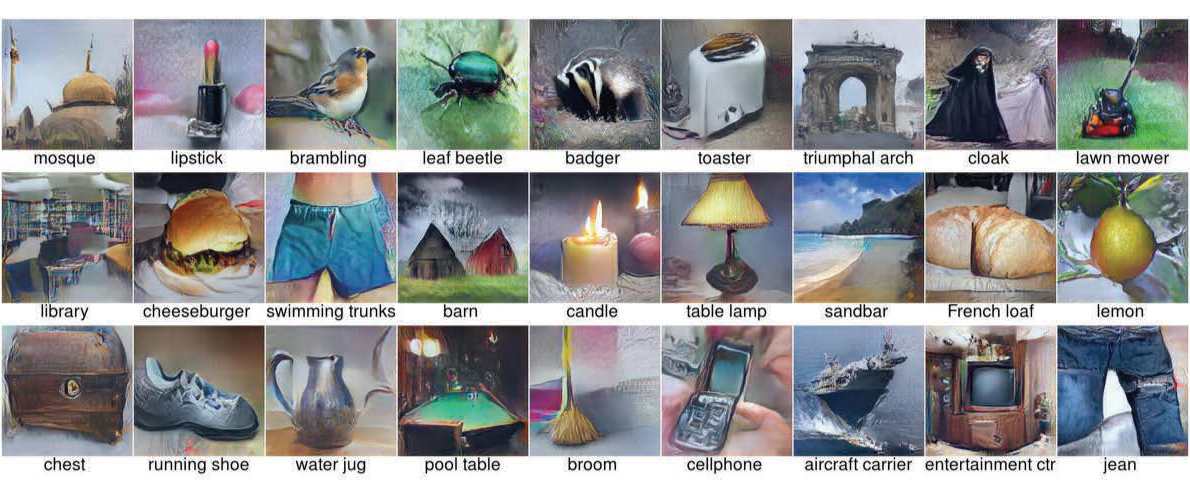}
\end{center}
	\caption{To explain what a black box classifier network comprehends as a class member being \cite{nguyen2016synthesizing} made synthetic prototype images that look real. They were created by a deep generator network and classified by the black box neural network.}
	\label{fig:deepgen}
\end{figure}

Another technique was realized by \cite{simonyan2013deep}. To find \textbf{prototype class members}, they created input images that have the highest probability to be predicted as certain classes of a trained CNN. Their tools are Taylor series, based on partial derivatives to display input sensitivities in images.
A few years later \cite{nguyen2016synthesizing} developed this idea further by synthesizing the preferred inputs for neurons in neural networks via deep generator networks for activation maximizing. The first algorithm is the generator and creates synthetic prototype class members that look real. The second algorithm is the black box classifier of the artificial image whose classification probability should be maximized. To view the prototype images, see Figure \ref{fig:deepgen}.
Another related derivative-based method is \textbf{DeepLift} \cite{shrikumar2017learning}. It propagates activation differences instead of gradients through the network. Partial derivatives do not explain a single decision but point to what change in the image could make a change in the prediction.

\begin{figure}[H]
	\begin{center}
\includegraphics[width=0.4\linewidth]{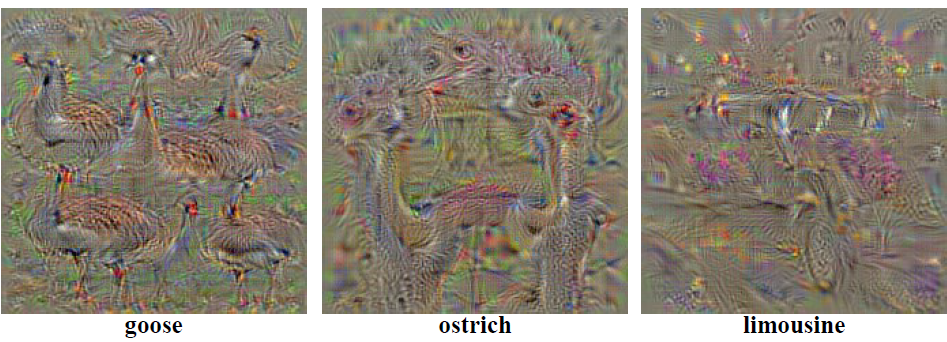}
\end{center}
	\caption{Deep inside convolutional networks \cite{simonyan2013deep}: created input images that have the highest probability to be predicted as certain classes of a trained CNN. Here one can see the created prototypes of the class goose, ostrich, limousine (left to right).}
	\label{fig:deepins}
\end{figure}

\begin{figure}[H]
	\begin{center}
\includegraphics[width=0.6\linewidth]{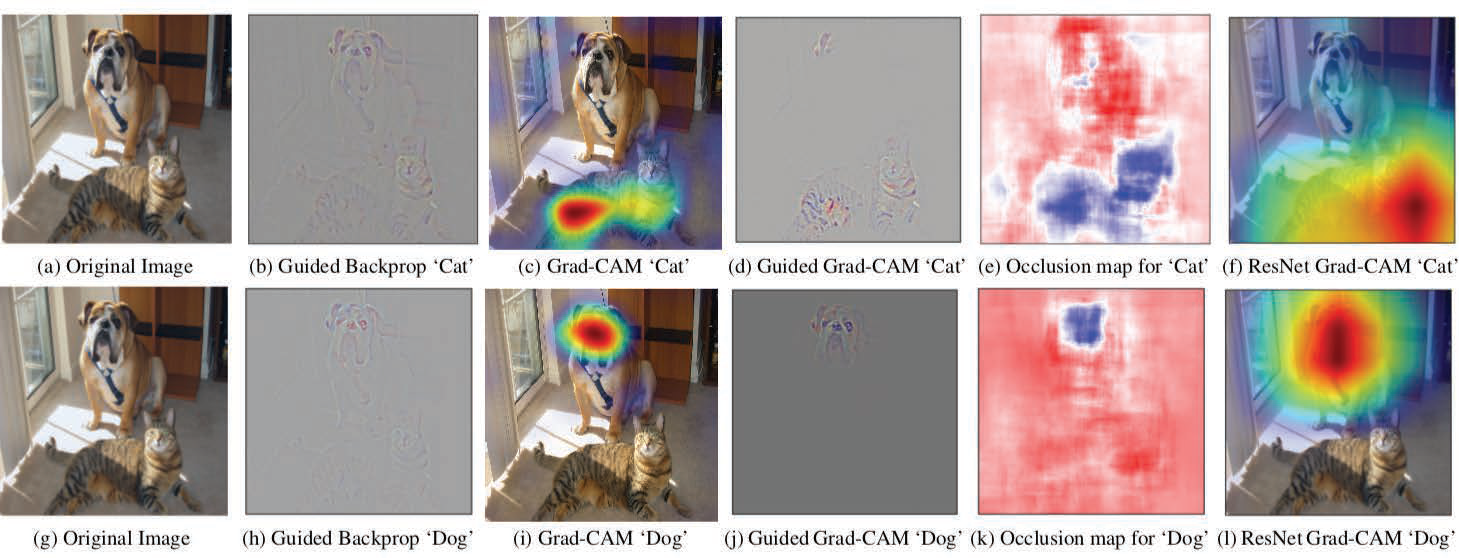}
\end{center}
	\caption{Gradient-weighted Class Activation Mapping (Grad-CAM) \cite{selvaraju2017grad} explained the outcome decision cat or dog, respectively, of an input image using the gradient information to understand the importance of each neuron in the last convolutional layer of the CNN.}  
	\label{fig:gradcam}
\end{figure}

\cite{zhou2016learning} showed that some convolutional layers behave as unsupervised  object detectors. They use global average pooling and create heat maps of a pre-softmax layer that point out the regions of an image which is responsible for a prediction. The method is called Class Activation Mapping (\textbf{CAM}). 
Upon this was created Gradient-weighted Class Activation Mapping (\textbf{Grad-CAM}) \cite{selvaraju2017grad}, see Figure \ref{fig:gradcam}, which is applicable to several CNN model-families, classification, image captioning, visual question answering, reinforcement learning, or re-training. A outcome decision can be explained by Grad-CAM through using the gradient information to understand the importance of each neuron in the last convolutional layer of the CNN. The Grad-CAM localizations are combined with existing high-resolution visualizations to obtain high-resolution class-discriminative guided Grad-CAM visualizations as saliency masks.
On the methods CAM and Grad-CAM were built \textbf{Grad-CAM++} \cite{gradcam++2017} that gives human interpretable visual explanations of CNN-based predictions of multiple tasks like classification, image captioning, or action recognition. Grad-CAM++ explains by regarding occurrences of multiple object instances in an image, combining the positive partial derivatives of feature maps of a convolutional rear layer with a weighted special class score.

\begin{figure}[H]
	\begin{center}
\includegraphics[width=0.4\linewidth]{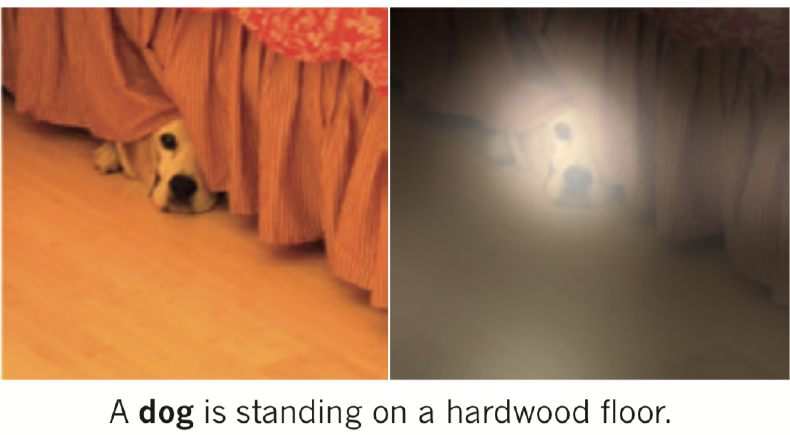}
\end{center}
	\caption{Deep Learning, \cite{lecun2015deep}: This method highlights the region of an image (dog) which is most important for the part ``dog" of the predicted output ``A dog is standing on a hardwood floor" of a trained CNN.}
	\label{fig:lecun}
\end{figure}

To mark the most responsible pixels or areas of pixels of an image for a special class prediction it is a promising idea to increase human understanding. The approach of \cite{lecun2015deep} focuses on single words of a caption generated by a RNN and highlights the region of the image which is most important for this word, see Figure \ref{fig:lecun}.

Much more general is Local Interpretable Model-agnostic Explanations (\textbf{LIME}) presented by \cite{ribeiro2016should}, which can explain the predictions of any agnostic black box classifier and any data, see Figure \ref{fig:lime}. It is a post-hoc, local model, interpretable and model-agnostic. LIME focuses on feature importance and gives outcome explanations: it highlights the super-pixels of the regions of the input image or the words from a text or tabular which are relevant for the given prediction. The disadvantage of this popular explainer is that it is itself a black box and in addition to this, \cite{anonymous2020on} pointed to the poor performance of LIME during their proposed evaluation metrics correctness, consistency, and confidence in comparison to the other regarded explainers Grad-CAM, Smooth-Grad and IG (described in the following).
 Furthermore, one can explain only images which can be split in super-pixels. The authors do not describe how to explain video object detection or segmentation networks.
In addition to this, the Submodular Pick (\textbf{SP-LIME}) algorithm judges whether you can trust the whole model or not.  It selects a picked diverse set of representative instances with LIME explanations via submodular optimization. The user should evaluate the black box by regarding the feature words of the selected instances. With the knowledge it is also possible to improve a bad model. SP-LIME was researched with text data, but the authors claim that it can be transferred to any data type models.

\begin{figure}[H]
	\begin{center}
\includegraphics[width=0.6\linewidth]{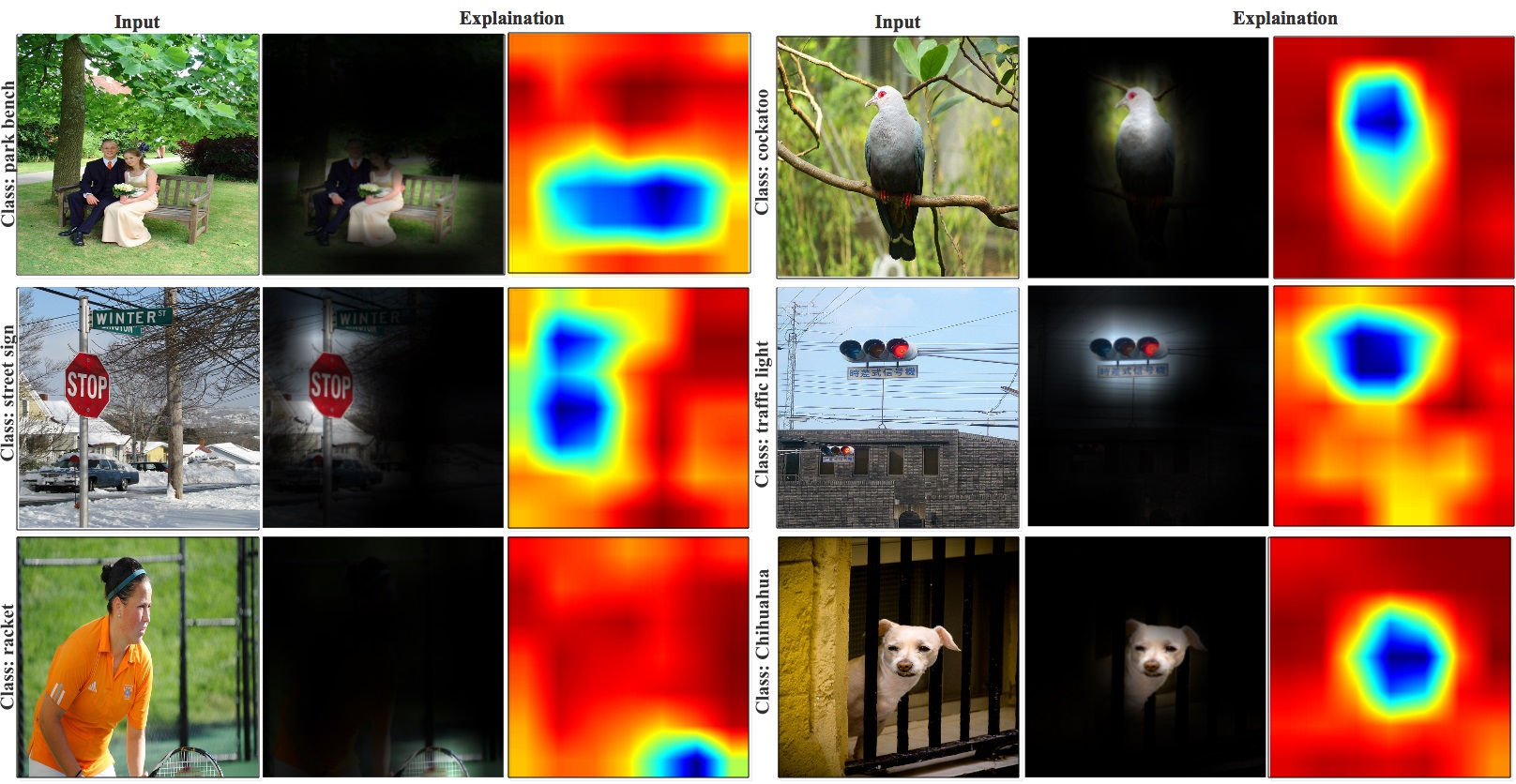}
\end{center}
	\caption{
	Deep Visual Explanation \cite{babiker2017introduction} highlights the most discriminative region in an image of six examples (park bench, cockatoo, street sign, traffic light, racket, chihuahua) to explain the decision made by the VGG-16.}
	\label{fig:hal}
\end{figure}

Another approach that focuses on the most discriminative region in an image to explain an automatic  decision is	Deep Visual Explanation (\textbf{DVE}) \cite{babiker2017introduction}, see Figure \ref{fig:hal}. They were inspired from CAM and Grad-CAM and tested the explanator to randomly chosen images from the COCO dataset \cite{lin2014microsoft}, applied to the pre-trained neural network VGG-16 using Kullback Leibler (KL)-divergence \cite{babiker2017using}. They captured the discriminative areas of the input image by considering the activation of high and low spatial scales in Fourier space.

With their conditional multivariate model Prediction Difference Analysis (\textbf{PDA}) \cite{zintgraf2017visualizing} have concentrated on explaining visualizations of natural and medical images in classification tasks. Their goal has been to improve and interpret DNNs. Their technique is based on the uni-variate approach of \cite{robnik2008explaining} and the idea that the relevance of an input feature with respect to a class can be estimated by measuring how the prediction changes if the feature is removed. Zintgraf et al. remove several features at one time using their knowledge about images by strategically choosing patches of connected pixels as feature sets. Instead of going through all individual pixels, they regard all patches of a special size implemented in a sliding window fashion. They visualize the effects of different window sizes and marginal versus conditional sampling and display feature maps of different hidden layers and top scoring classes.

\cite{smilkov2017smoothgrad} described \textbf{Smooth-Grad} that reduces visual noise and hence, improves visual explanations how a DNN is making a classification decision. Comparing their work to several gradient-based sensitivity map methods like LRP, DeepLift and Integrated Gradients (\textbf{IG}) \cite{sundararajan2017axiomatic} that estimate the global importance of each pixel and create saliency maps, shows that Smooth-Grad focuses on local sensitivity and calculates averaging maps with a smoothing effect made from several small perturbations of a input image. The effect is enhanced by further training with these noisy images and finally reaches an impact on the quality of sensitivity maps by sharpening them. The work of \cite{anonymous2020on} evaluated the explainers LIME, Grad-CAM, Smooth-Grad and IG with regard to the properties correctness, consistency, and confidence and came to the result that Grad-CAM often performs better than others.\\

\begin{figure}[H]
	\begin{center}
\includegraphics[width=0.4\linewidth]{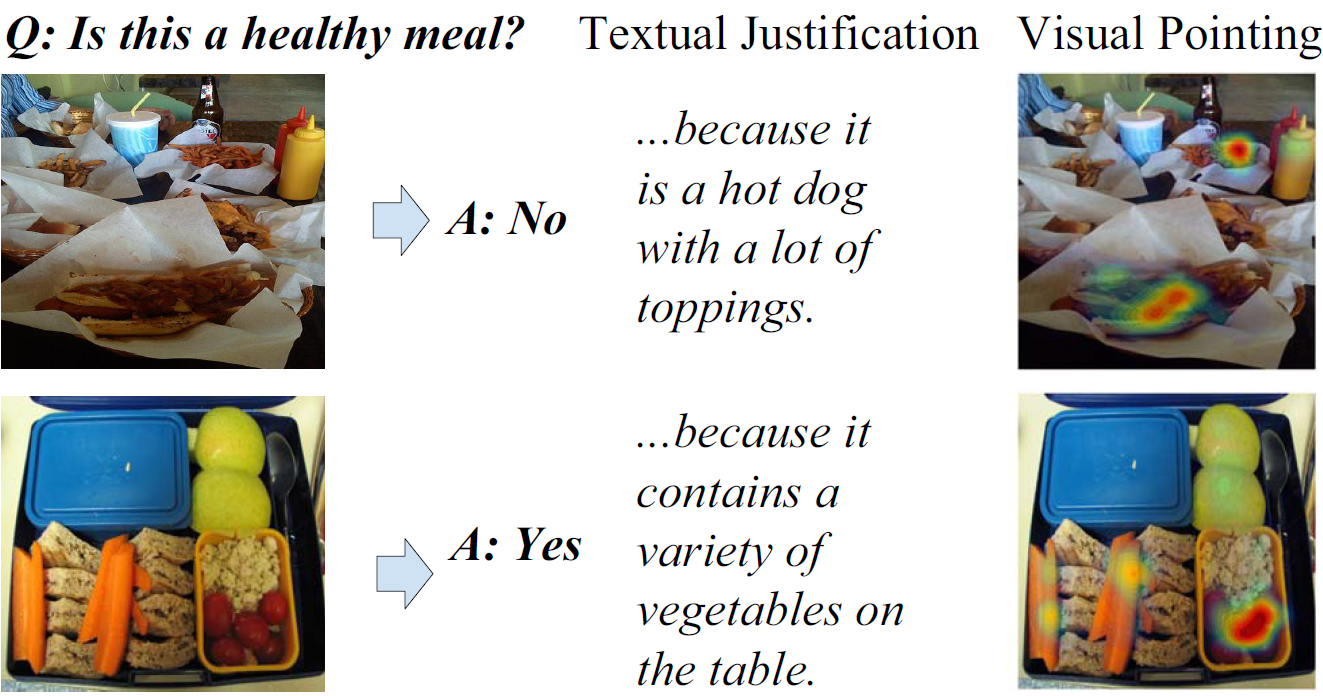}
\end{center}
	\caption{Multimodal Explanation (\textbf{ME}) \cite{huk2018multimodal}  explains by two types of
justifications of visual question answering task: The example shows two images with food and the question is, if they contain healthy meals or not. The explanations of the answers ``Yes" or ``No" are given textual in justifying in a sentence and visual in pointing out the most responsible ares of the image.}
	\label{fig:park}
\end{figure}

To improve and expend understanding Multimodal Explanation (\textbf{ME}) \cite{huk2018multimodal}, a local, post-hoc approach gave visual and textual justifications of predictions with the help of two novel explanation datasets through crowd sourcing. The employed tasks were classification decision for activity recognition and visual question answering, see Figure \ref{fig:park}. The visual explanation was created by an attention mechanism which conveys knowledge about what region of the image is important for the decision. This explanation guides to generate the textual justification out of a LSTM feature that is a prediction of a classification problem over all possible justifications.

\begin{figure}[H]
	\begin{center}
\includegraphics[width=0.6\linewidth]{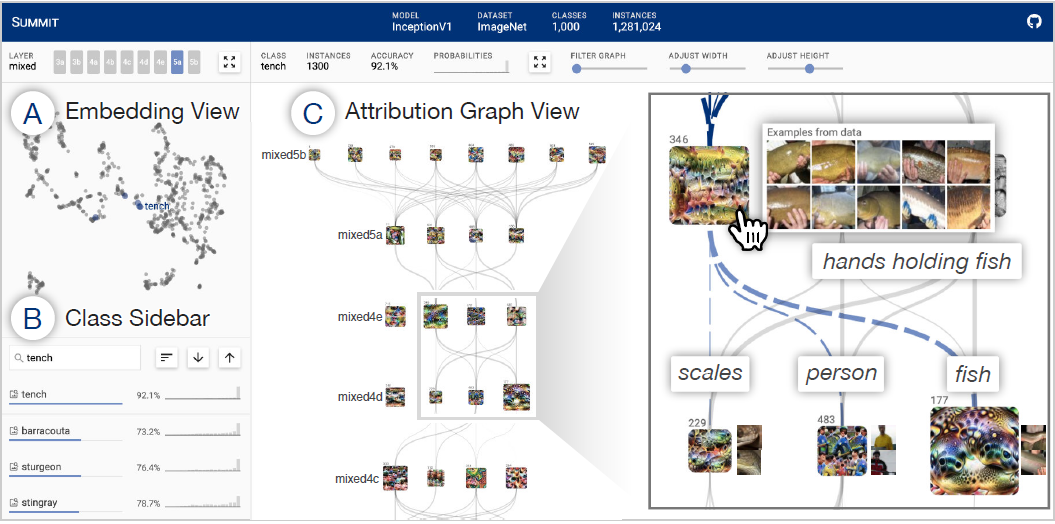}
\end{center}
	\caption{Summit \cite{hohman2019summit} visualizes what features a Deep Learning model has learned and how those features are connected to make predictions. The Embedding View (A) shows which classes are related to each other, Class Sidebar (B) is linked to the embedding view listing all classes sorted in several ways, and the Attribution Graph (C) summarizes  crucial neuron associations and substructures that contribute to a model’s prediction.}
	\label{fig:summit}
\end{figure}
\begin{figure}[H]
	\begin{center}
\includegraphics[width=0.4\linewidth]{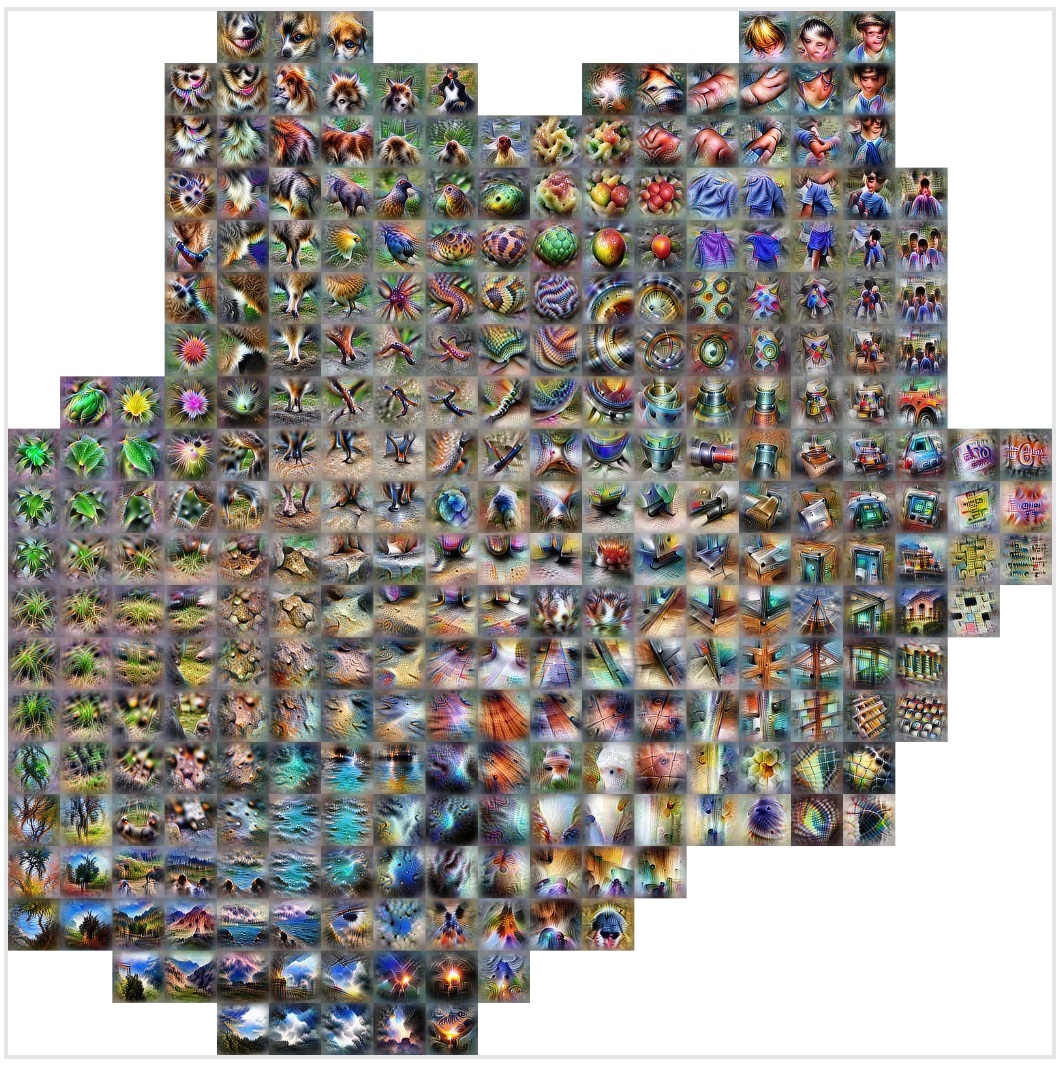}
\end{center}
	\caption{
	Activation atlases with 100,000 activations, \cite{carter2019activation}.}
	\label{fig:act}
\end{figure}

In this year a new and extensive visualized approach was created by \cite{hohman2019summit}, see Figure \ref{fig:summit}, showing what features a Deep Learning model has learned and how those features interact to make predictions. Their model is called \textbf{Summit} and combines two scalable tools: (1) activation aggregation discovers important neurons, and (2) neuron-influence aggregation identifies relationships among such neurons. An attribution graph that reveals and summarizes crucial neuron associations and substructures that contribute to a model’s outcomes is created. Summit combines famous methods like computing synthetic prototypes of features and showing examples from the dataset that maximize special neurons of different layers. Deeper in the graph is examined how the low-level-features combine to create high-level features. Also novel is exploiting neural networks with \textbf{activation atlases} \cite{carter2019activation}, see Figure \ref{fig:act}. This method uses feature inversion to visualize millions of activations from an image classification network to create an explorable activation atlas of features the network has learned. Their approach is able to reveal visual abstractions within a model and even high-level misunderstandings in a model that can be exploited. Activation atlases is a novel way to peer into convolutional vision networks and represents a global, hierarchical, and human-interpretable overview of concepts within the hidden layers. 

Just in the last months the importance of interpretability was growing, that is why there appeared several single studies, investigating the contribution of some aspects of a neural network like the impact of color \cite{buhrmester2019evaluating}, texture \cite{geirhos2018imagenet} etc. without explaining a whole model extensive, however, which all contribute to a deeper understanding. \\
In Table \ref{tab} we give an overview about the presented explainers of DNNs, sorted by data and year. The main techniques and properties are mentioned for a short comparison. The property model-agnostic is abbreviated with agn.

\begin{table}[H]
\caption{Overview about some explainers for DNNs. Ordered by model or paper name, reference (author and year), data type, main method and main properties.} 
\label{tab}
\centering
\begin{tabular}{c c c c c c c} 
\toprule \textbf{Model} &\textbf{authors}& \textbf{data} & \textbf{method} &\textbf{properties} \\
\midrule
Deep Inside&\cite{simonyan2013deep}&image&saliency mask&local, post-hoc\\
Deconvnet&\cite{zeiler2014visualizing} &image&gradients&global, post-hoc\\
All-CNN&\cite{springenberg2015striving}&image&gradients&global, post-hoc\\
Deep Visualization& \cite{yosinski2015understanding}&image&neurons activation&global, ante-hoc\\
Deep Learning&\cite{lecun2015deep}&image&visualization&local, post-hoc\\
Show, attend, tell& \cite{xu2015show}&image&saliency mask&local, ante-hoc\\
LRP&\cite{bach2015pixel}&image&decomposition&local, ante-hoc\\
CAM& \cite{zhou2016learning}&image&saliency mask&local, post-hoc\\
Deep Generator&\cite{nguyen2016synthesizing}&image&gradients, prototype&local, ante-hoc\\
Interpretable DNNs&\cite{sturm2016interpretable}&image&saliency map&local, ante-hoc\\
VBP&\cite{bojarski2016visualbackprop}&image&saliency maps&local, post-hoc\\
DTD&\cite{montavon2017explaining}&image&decomposition&local, post-hoc\\
Meaningful&\cite{fong2017interpretable} &image&saliency mask&local, post-hoc, agn.\\
PDA&\cite{zintgraf2017visualizing}&image&feature importance&local, ante-hoc\\
DVE&\cite{babiker2017using}&image&visualization&local, post-hoc\\
Grad-CAM&\cite{selvaraju2017grad}&image&saliency mask&local, post-hoc\\
Grad-CAM++ & \cite{gradcam++2017}&image&saliency mask&local, post-hoc\\
Smooth-Grad&\cite{smilkov2017smoothgrad}&image&sensitivity analysis&local, ante-hoc\\
ME& \cite{huk2018multimodal}&image&visualization&local, post-hoc\\
Summit&\cite{hohman2019summit}&image&visualization&local, ante-hoc\\
Activation atlases&\cite{carter2019activation}&image&visualization&local, ante-hoc\\
SP-LIME&\cite{ribeiro2016should}&text&feature importance&local, post-hoc\\
Rationalizing& \cite{lei2016rationalizing}&text&saliency mask&local, ante-hoc\\
Generate reviews&\cite{radford2017learning}&text&neurons activation&global, ante-hoc\\
BRL&\cite{letham2015interpretable}&tabular&decision tree&global, ante-hoc\\
TreeView& \cite{thiagarajan2016treeview}&tabular&decision tree&global, ante-hoc \\
IP&\cite{shwartz2017opening}&tabular&neurons activation&global, ante-hoc\\
KT&\cite{fu1994rule}&any&rule extraction&global, ante-hoc\\
Decicion Tree&\cite{friedman2001elements}&any&rule extraction&global, ante-hoc, agn.\\
CIE&\cite{bottou2013counterfactual}&any&feature importance&local, post-hoc\\
DeepRed&\cite{zilke2016deepred}&any&rule extraction&global, ante-hoc\\
LIME&\cite{ribeiro2016should}&any&feature importance&local, post-hoc, agn.\\
NES&\cite{turner2016model}&any&rule extraction&local, ante-hoc\\
BETA&  \cite{lakkaraju2016interpretable}&any&decision tree&global, ante-hoc\\
PALM&\cite{krishnan2017palm}&any&decision tree&global, ante-hoc\\
DeepLift&\cite{shrikumar2017learning}&any&feature importance&local, ante-hoc\\
IG& \cite{sundararajan2017axiomatic}&any&sensitivity analysis&global, ante-hoc\\
RETAIN&\cite{choi2016retain}&EHR&Reverse Time Atten.&global, ante-hoc\\
\bottomrule
\end{tabular}
\end{table}

\subsection{Analysis of understanding and explaining methods}
\label{sec:selected2}
We still want to go into studies on the general analysis of explainability and machine understanding.
To reach a better understanding of multilayer neural network \cite{agrawal2014analyzing} analyzed pre-training and fine-tuning of several classification and object recognition tasks. They found that some CNN-learned features are grandmother-cell-like, that are cells in the human brain which only respond to very specific and complex visual stimuli, such as the face of one's grand-mother \cite{quiroga2005invariant} and that a longer pre-training significantly improves the performance.
 \cite{samek2016evaluating} investigated different methods to compute heat maps in Computer Vision application. They concluded that layer-wise relevance propagation algorithms (e.g. LRP) were qualitatively and quantitatively superior in explaining what made a DNN arrive at a particular classification decision to the sensitivity-based approaches or the deconvolution methods \cite{zeiler2014visualizing}. The inferior methods were much noisier and less suitable for identifying the most important regions with respect to the classification task. Their work did not give an answer how to make a more detailed analysis of prioritization of image regions or even how to quantify the heatmap quality. 
\cite{kindermans2017learning} criticized the explaining methods Deconvnet, guided Backpropagation and LRP not to produce the theoretically correct explanation for a linear model and their contributions to understanding were scarce.
Based on an analysis of linear models, see also \cite{haufe2014interpretation}, \cite{montavon2017explaining} they propose a generalization that yielded the two neuron-wise explanation techniques \textbf{PatternNet} (for signal visualization) and \textbf{PatternAttribution} (for decomposition methods) by taking the data distribution into account. They demonstrated that their methods were sound and constitute a theoretical, qualitative and quantitative improvement towards explaining and understanding DNNs.

\textbf{SHAP} (SHapley Additive exPlanations) \cite{lundberg2017unified} has been developed to interpret a models prediction by additive feature attribution methods. It unifies six previously existing explanations and assigns each feature an importance value for a particular prediction. Its new components include: (1) the identification of a new class of additive feature importance measures, and (2) theoretical results showing there is a unique solution in this class with a set of desirable properties. The framework SHAP reaches a higher performance and a better consistency with human intuition than previous approaches in this tasks, as the authors claim.

\cite{bau2017network} made an approach to quantify interpretability of Deep Visual Representation by proposing a general framework called \textbf{Network Dissection} by evaluating the alignment between individual hidden units and a set of semantic concepts. The proposal raises how to detect and evaluate disentangled representations and investigates if there is a coherence between hidden units and a special alignment of feature space. Also the influence of training conditions on entanglement of explanations is regarded and confirmed to have a significant effect of the representation learned by the hidden units. Their framework Network Dissection comes to the conclusion that interpretability is not an axis-independent phenomenon, which is consistent with the hypothesis that interpretable units indicate a partially disentangled representation.

With an overview of interpretability of Machine Learning \cite{gilpin2018explaining} try to explain explanations. They define some key terms and review a number of approaches towards classical explainable AI systems also focusing on Deep Learning tasks. Furthermore, they investigate the role of single layers, individual units and representation vectors in explanation of deep network representations. Finally, they present a taxonomy that examines what is being explained by these explanations. They summarize that it is not obvious what the best purpose or type of explanation metric is and should be and give the advice to combine explaining ideas from different fields.

Another approach towards understanding is to evaluate the human-interpretability of explanation \cite{narayanan2018humans}. They investigated the consistence of output of a ML-system with its input and the supposed rationale. Therefor they carried out user-studies to identify what kind of increases in complexity have the most dominant effect on the time humans need to take for verification the rationale, and which seem to be more insensitive. Their study quantifies what kind of explanation makes them to be most understandable by humans. As a main result they found out that in general, greater complexity results in higher response times and lower satisfaction.

Even simple interpretable explainers do mostly not quantify if the user can trust them. A study on trust in black box models and post-hoc explanations, \cite{el2019study} provides the problems in main problems in literature and their kind of black box systems. They evaluate three different explanation approaches: (1) based on the users’ initial trust, (2) the users’ trust in the provided explanation of three different post-hoc explanator approaches and (3) the established trust in the black box by a within-subject design study. The participants where asked if they trust that a special algorithm works well in the real world, if they suppose it to be able to distinguish between the classes and why. The results of their work led to the conclusion, that although the black box prediction model and explanation are independent of each other, trusting the explanation approach is an important aspect of trusting the prediction model.

\begin{figure}[H]
	\begin{center}
\includegraphics[width=0.6\linewidth]{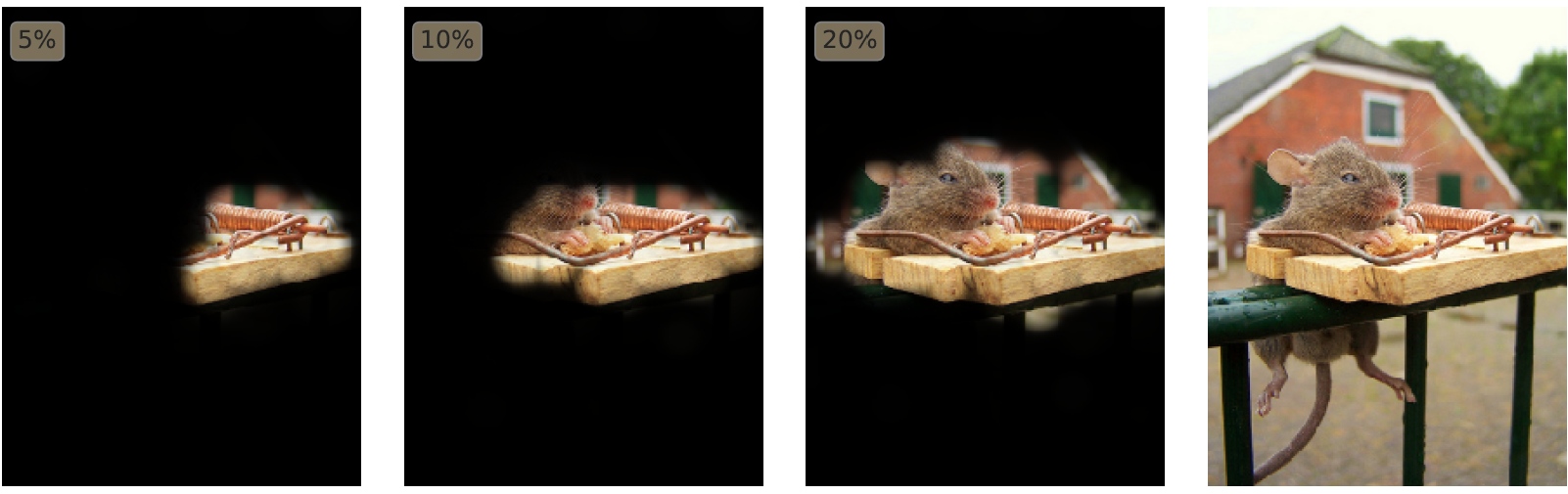}
\end{center}
	\caption{
	Extremal perturbations \cite{fongunderstanding}: The example shows the regions of an image
(boxed) that maximally affect the activation of a certain neuron in a DNN (“mousetrap” class score). For clarity, the masked regions are blacked out. In practice, the network sees blurred regions.}
	\label{fig:experturb}
\end{figure}

A discussion of some existing approaches to perturbation analysis is done in the study of \cite{fongunderstanding}, Figure \ref{fig:experturb}. Their work, based on \cite{fong2017interpretable}, found adversarial effect of perturbations on the network’s output. Extremal perturbations are regions of an input image that maximally affect the activation of a certain neuron in a DNN. Measuring the effects of perturbations of the image is an important family of attribution methods. Attribution aims at characterising the response of DNN by looking at which parts of the network’s input are the most responsible ones for determining its prediction, which is mostly done by several kinds of backpropagation. Fong et al. investigate their effect as a function of the area. In particular, they have visualized the difference between perturbed and unperturbed activations using a representation inversion technique. They introduced TorchRay \cite{torchray}, a PyTorch interpretability library.

\subsection{Open problems of understanding DNNs}
\label{sec:open}

When summarizing the functionality of explainers, one notices that some facts are hard to measure:
First, the time, which is needed to understand the decision is difficult to take. Local working explainers can deliver a root case for each prediction, but how many examples are necessary to look at, to be sure, that all results and thereby the black box is faithful? In addition to this, the model complexity of several explainers is different. The complexity is often calculated as an opposed term to interpretability. Complexity of a black box can be expressed for instance as the number of non-zero weights at neural networks or the depth of trees for decision. But the complexity of the explanation could depend of the complexity of the black box.\\
More work must be done also in data exploration: Interpretable data in the mentioned papers are mainly images, texts and tabular data, which is all easily interpretable data for humans. Missing is general data like vectors, matrices, or complex spatio-temporal numbers. Of course they have to be transformed before analyzing to be understandable from our brains. Sequences, networks etc. could be an input in a black box, but until now, such models are not explained.\\
There is no agreement how to quantify the grade of explanation. This is an open problem. Some metrics to evaluate explainers are proposed, e.g. \cite{anonymous2020on}, but unfortunately, many of them tend to suffer from major drawbacks like computational cost \cite{hooker2018evaluating} or simply focusing on one desirable attribute of a good explainer \cite{yeh2019sensitive}.
But a definition with properties of a DNN model like reliability, soundness, completeness, compactness, comprehensibility, and the knowledge of breaking points of an algorithm are still missing. However, there is a need in focusing on uniform definitions. Important to regard could be the robustness of model or input data which indicates how robust the system is in small changes of the architecture or the test date. To compare models also reliability and fairness need to be quantified. Our further work will be done here.

\section{Conclusion}
\label{sec:conclusion}

In this paper we give reasons why it is necessary to comprehend black box Deep Neural Networks: Adverserial attacks that are not understandable by humans pose a serious threat to the robustness of DNNs.  Furthermore, we expound why models learn prejudices through contaminated training data, hence, through their widespread application they are responsible for grown unfairness. On the other hand, novel laws demand for the right of explanation for users of intelligent decision-making systems. One first solution is the development of explainers. We give a taxonomic introduction in the main definitions, properties, and mechanisms of explaining systems. Unlike others -- to our best knowledge -- we focus only on state-of-the-art explainers for DNNs, especially on the area Computer Vision and compare their similar or different methods and representations of explanations. We differ the explainers with regard to technical mechanisms, application, data, and  pros and cons. Finally, we introduce surveys and studies, that analyze or evaluate explaining systems and try to quantify machine understanding in general.
In addition to this, summarizing open problems with an outlook of further ideas and specifying some missing definitions of explainers makes this work useful for Computer Vision scientists. 



\reftitle{References}


\externalbibliography{yes}
\bibliography{intbib}





\end{document}